\documentclass[twoside,11pt]{article}

\usepackage{blindtext}

\usepackage[preprint, nohyperref]{jmlr2e}
\usepackage{booktabs}

\usepackage{xcolor}
\definecolor{deep_blue}{RGB}{31, 119, 180}
\definecolor{deep_orange}{RGB}{255, 127, 14}
\definecolor{deep_green}{RGB}{44, 160, 44}
\definecolor{deep_red}{RGB}{214, 39, 40}
\definecolor{deep_purple}{RGB}{148, 103, 189}
\definecolor{deep_brown}{RGB}{140, 86, 75}
\definecolor{deep_pink}{RGB}{227, 119, 194}

\usepackage[colorlinks,linktoc=all]{hyperref}
\hypersetup{
  linkcolor=deep_pink,    
  citecolor=deep_green,   
  urlcolor=deep_red       
}

\usepackage{lastpage}
\jmlrheading{23}{2025}{1-\pageref{LastPage}}{}{}{21-0000}{Francesco Martinuzzi}

\ShortHeadings{Unified Implementations of Recurrent Neural Networks}{Martinuzzi}
\firstpageno{1}

\begin{document}

\title{Unified Implementations of Recurrent Neural Networks in Multiple Deep Learning Frameworks}

\author{\name Francesco Martinuzzi \email martinuzzi@pks.mpg.de \\
       \addr Max Planck Institute for the Physics of Complex Systems, \\
       Dresden, Germany}

\editor{}

\maketitle

\begin{abstract}
    Recurrent neural networks (RNNs) are a cornerstone of sequence modeling across various scientific and industrial applications. Owing to their versatility, numerous RNN variants have been proposed over the past decade, aiming to improve the modeling of long-term dependencies and to address challenges such as vanishing and exploding gradients.
    However, no central library is available to test these variations, and reimplementing diverse architectures can be time-consuming and error-prone, limiting reproducibility and exploration.
    Here, we introduce three open-source libraries in Julia and Python that centralize numerous recurrent cell implementations and higher-level recurrent architectures.
    \texttt{torchrecurrent}, \texttt{RecurrentLayers.jl}, and \texttt{LuxRecurrentLayers.jl} offer a consistent framework for constructing and extending RNN models, providing built-in mechanisms for customization and experimentation. All packages are available under the MIT license and actively maintained on GitHub.
\end{abstract}


\section{\label{rl:sec:intro} Introduction}
    Sequential data, such as time series, are pervasive across multiple scientific disciplines \citep{manning1999foundations, goldberger2000physio, kantz2003nonlinear}. With the increasing availability of large datasets and computational resources,  modern modeling approaches increasingly rely on deep learning (DL) models \citep{lngkvist2014review, lim2021time}. Among DL architectures, recurrent neural networks (RNNs) stand out due to their explicit modeling of temporal dependencies \citep{jordan1986attractor, elman1990finding}. Thanks to their embedded temporal inductive bias, RNNs have found applications in diverse fields, including Earth sciences \citep{kratzert2018rainfall, arcomano2020machine, martinuzzi2024learning}, speech and audio processing \citep{graves2013speech, hannun2014deep}, neuroscience \citep{yang2019task, durstewitz2023reconstructing}, and dynamical system reconstruction \citep{pathak2017using, vlachas2020backpropagation}. This wide range of applications has, in turn, motivated the development of numerous RNN variants, each introducing unique design principles. Examples include architectures with gating mechanisms \citep{hochreiter1997long, cho2014learning, zhou2016minimal}, physics-inspired dynamics \citep{rusch2021coupled, rusch2022long}, and geometric constraints \citep{arjovsky2016unitory, chang2018antisymmetricrnn, kerg2019nonnormal}.

    Despite the diversity of proposed architectures, only a small subset of RNNs are supported by mainstream DL libraries. Code accompanying individual papers may exist, but it is often hardcoded for specific tasks, not continuously maintained, or incompatible with current frameworks. Additionally, a central repository of unified implementations of RNNs is lacking in any DL framework. The absence of centralized, streamlined, and up-to-date implementations restricts researchers’ ability to test or extend alternative designs, concentrating empirical work on a few canonical architectures. In fact, virtually all the mainstream DL libraries limit their offering to the Elman RNN \citep{elman1990finding}, the long short-term memory (LSTM; \citealp{hochreiter1997long}), and the gated recurrent unit (GRU; \citealp{cho2014learning}).

    To enable easier exploration of alternative RNN designs and increase the number of implementations available to researchers, we introduce three open-source libraries implementing a wide range of recurrent architectures: \textbf{\texttt{torchrecurrent}} (based on \texttt{PyTorch} \citep{paszke2019pytorch}), \textbf{\texttt{RecurrentLayers.jl}} (based on \texttt{Flux.jl} \citep{innes2018flux}), and \texttt{\textbf{LuxRecurrentLayers.jl}} (based on \texttt{Lux.jl} \citep{pal2023lux}). Collectively, we refer to these as \texttt{recurrent}. Each library provides unified primitives for constructing RNNs within its respective framework, along with multiple implementations of recurrent cells and layers. Models from \texttt{recurrent} can be invoked using the same interface as their native counterparts, allowing easy integration into existing workflows. At the same time, the layers in \texttt{recurrent} provide additional configuration options to support customization and exploration of novel RNN architectures. By standardizing these implementations, \texttt{recurrent} lowers the barrier to experimentation and accelerates the development of new recurrent models.

    \section{\label{rl:sec:overview} Software Overview}

        \begin{figure}[ht]
            \centering
            \includegraphics[width=0.85\textwidth]{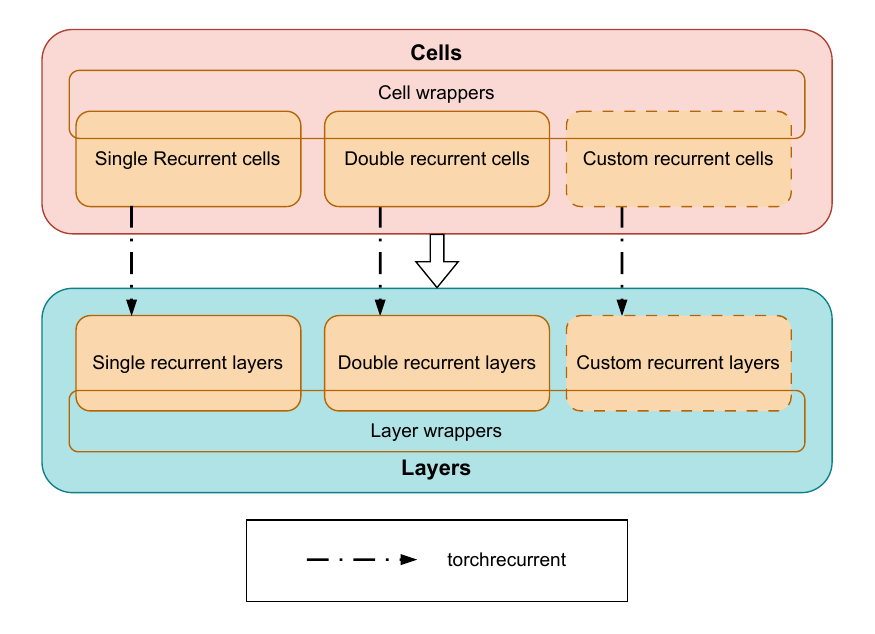}
            \caption{\textbf{Structure of recurrent components in the \texttt{recurrent} libraries.} Each library defines recurrent \emph{cells} and corresponding \emph{layers}, which can be extended using \emph{wrappers}. Single, double, and custom recurrent architectures share a unified interface, while framework-specific differences (e.g., in \texttt{torchrecurrent}) are indicated with dashed arrows.}
            \label{rl:fig:structure}
        \end{figure}

        \paragraph{\label{rl:par:install} Installation}
            All three libraries are registered in the main software repositories of their respective languages. The Julia packages are available in the General Registry\footnote{\href{https://github.com/JuliaRegistries/General/tree/master/R/RecurrentLayers}{github.com/JuliaRegistries/General/tree/master/R/RecurrentLayers} and \href{https://github.com/JuliaRegistries/General/tree/master/L/LuxRecurrentLayers}{github.com/JuliaRegistries/General/tree/master/L/LuxRecurrentLayers}}, while \texttt{torchrecurrent} can be installed directly from PyPI\footnote{\href{https://pypi.org/project/torchrecurrent/}{pypi.org/project/torchrecurrent/}}. Because the primary dependency of each library is its corresponding deep learning framework, no additional installation steps are required for components such as graphical processing unit (GPU) support.

        \paragraph{\label{rl:par:models} Models}
            The libraries implement a diverse set of recurrent architectures, organized into three levels of abstraction: \emph{cells}, which represent single computational units; \emph{layers}, which apply these computations across full sequences; and \emph{wrappers}, which extend either cells or layers with additional behaviors. Wrapped components retain the same interface as their unwrapped counterparts, allowing them to be used interchangeably within existing workflows. Currently, \texttt{recurrent} includes approximately 30 recurrent cells, each with corresponding layers and optional wrappers. The available cells span a wide range of designs, including gated architectures \citep{kusupati2018frastgrnn, ravanelli2018light, landi2021working}, physics-inspired formulations \citep{rusch2021coupled, rusch2022long}, and models that use alternative integration schemes \citep{krause2017multiplicative, li2018independently}.

            While the libraries currently provide largely overlapping model implementations, the development of each library progresses independently. Each library may gradually specialize in a specific niche, shaped by the interests of its user community. For instance, given the strong focus on scientific machine learning within the Julia ecosystem \citep{rackauckas2019diffeqflux, zubov2021neuralpde, martinuzzi2022reservoircomputing}, it is likely that \texttt{LuxRecurrentLayers.jl} will evolve toward providing implementations of continuous-time RNNs \citep{brouwer2019gruodebayes, lechner2020learning}. Conversely, because of the widespread use of \texttt{PyTorch} for large-scale models, \texttt{torchrecurrent} may expand to include hybrid architectures based, for instance, on state-space models \citep{gu2022on, gu2022efficiently}.

        \paragraph{\label{rl:par:extensions} Modularity and Extensibility}
            In addition to pre-defined models, \texttt{recurrent} provides lightweight primitives to simplify the implementation of custom recurrent layers. By standardizing standard boilerplate code for recurrent models, such as parameter initialization, the libraries allow users to focus primarily on defining the forward computation. Once a custom cell is specified using the provided types and interfaces, it can automatically leverage higher-level utilities such as the layers and wrappers described above. Conversely, the packages' modularity allows users to apply custom wrappers to all the provided cells and layers. Figure~\ref{rl:fig:structure} illustrates the high-level structure that the packages follow. All models in the three libraries follow the identical function signatures as those of the underlying deep learning frameworks. This design choice ensures that new components can be easily integrated into existing workflows without introducing breaking changes.

    \section{\label{rl:sec:cq} Code Quality}

        The codebase for all three libraries is open source and hosted on GitHub \footnote{\href{https://github.com/}{github.com}}, under a permissive MIT license. \texttt{recurrent}'s codebase follows the standards for each language, with clear formatting guidelines, enforced through black in Python and JuliaFormatter in Julia. The libraries can be accessed at the following URLs:
        \begin{itemize}
            \item \texttt{torchrecurrent}: \href{https://github.com/MartinuzziFrancesco/torchrecurrent}{github.com/MartinuzziFrancesco/torchrecurrent}
            \item \texttt{RecurrentLayers.jl}: \href{https://github.com/MartinuzziFrancesco/RecurrentLayers.jl}{github.com/MartinuzziFrancesco/RecurrentLayers.jl}
            \item \texttt{LuxRecurrentLayers.jl}: \href{https://github.com/MartinuzziFrancesco/LuxRecurrentLayers.jl}{github.com/MartinuzziFrancesco/LuxRecurrentLayers.jl}
        \end{itemize}

        Each library undergoes rigorous testing through continuous integration using GitHub Actions. Tests are executed for every new package release as well as for each commit in pull requests. \texttt{LuxRecurrentLayers.jl} and \texttt{RecurrentLayers.jl} are tested against the long-term support, latest, and nightly versions of Julia, while \texttt{torchrecurrent} is tested across the range of Python versions supported by \texttt{PyTorch} v2.0.
        
        The libraries are accompanied by documentation hosted on GitHub. Currently, the documentation provides a detailed application programming interface (API) reference for each model in the library. Each model entry includes the forward equation implemented by that model, along with a detailed description of its arguments and keyword arguments. The documentation also lists all learnable parameters associated with the model.


    \section{\label{rl:sec:end} Conclusion and Outlook}

        \begin{sloppypar}
        In this paper, we introduce \texttt{torchrecurrent}, \texttt{RecurrentLayers.jl}, and \texttt{LuxRecurrentLayers.jl}, three libraries that provide implementations of a large number of RNNs across multiple DL frameworks in different programming languages. These libraries offer a unified approach to building RNNs, providing APIs that remain fully compatible with the underlying frameworks while extending their available functionality. These features make the presented libraries particularly useful for academic research.
        \end{sloppypar}
        Future development will focus on improving documentation and usability. Although all models are currently documented, their close alignment with the base deep learning libraries minimizes the need for extensive duplication. However, additional examples and use cases would help new users get started more easily. Another planned direction is establishing a centralized repository for RNN benchmarks built on top of these libraries. Finally, while the current focus is on model architectures, future extensions could include alternative training strategies and optimization schemes.

\acks{The author thanks Frank Loebe for supporting this work, and David Montero for his help in the visualization. The author acknowledges the financial support by the Federal Ministry of Research, Technology and Space of Germany and by Sächsische Staatsministerium für Wissenschaft, Kultur und Tourismus in the programme Center of Excellence for AI-research "Center for Scalable Data Analytics and Artificial Intelligence Dresden/Leipzig", project identification number: ScaDS.AI}

\vskip 0.2in
\bibliography{rl}

\end{document}